\documentclass[english,british]{article}
\usepackage[T1]{fontenc}
\usepackage{array}
\usepackage{multirow}
\usepackage{amsmath}
\usepackage{amsthm}
\usepackage{amssymb}
\usepackage{graphicx}

\makeatletter

\providecommand{\tabularnewline}{\\}

\theoremstyle{plain}
\newtheorem{thm}{\protect\theoremname}
\theoremstyle{plain}
\newtheorem{cor}{\protect\corollaryname}

\usepackage{aaai19}
\usepackage{times}
\usepackage{algorithm}
\usepackage{algorithmicx}
\usepackage{algpseudocode}

\@ifundefined{showcaptionsetup}{}{%
 \PassOptionsToPackage{caption=false}{subfig}}
\usepackage{subfig}
\makeatother

\usepackage{babel}
\addto\captionsbritish{\renewcommand{\corollaryname}{Corollary}}
\addto\captionsbritish{\renewcommand{\theoremname}{Theorem}}
\addto\captionsenglish{\renewcommand{\corollaryname}{Corollary}}
\addto\captionsenglish{\renewcommand{\theoremname}{Theorem}}
\providecommand{\corollaryname}{Corollary}
\providecommand{\theoremname}{Theorem}

\begin{document}

\title{Fast Hyperparameter Tuning using Bayesian Optimization with Directional
Derivatives}
\selectlanguage{english}%

\author{Tinu Theckel Joy, Santu Rana, Sunil Gupta, Svetha Venkatesh\\
Centre for Pattern Recognition and Data Analytics, Deakin University,
Australia}

\maketitle
\selectlanguage{british}%
%\pubnote{Submitted to AAAI 200x} 

\selectlanguage{english}%

\global\long\def\mF{\mathcal{F}}

\global\long\def\mA{\mathcal{A}}

\global\long\def\mH{\mathcal{H}}

\global\long\def\mX{\mathcal{X}}

\global\long\def\dist{d}

\global\long\def\HX{\entro\left(X\right)}
 \global\long\def\entropyX{\HX}

\global\long\def\HY{\entro\left(Y\right)}
 \global\long\def\entropyY{\HY}

\global\long\def\HXY{\entro\left(X,Y\right)}
 \global\long\def\entropyXY{\HXY}

\global\long\def\mutualXY{\mutual\left(X;Y\right)}
 \global\long\def\mutinfoXY{W\mutualXY}

\global\long\def\given{\mid}

\global\long\def\gv{\given}

\global\long\def\goto{\rightarrow}

\global\long\def\asgoto{\stackrel{a.s.}{\longrightarrow}}

\global\long\def\pgoto{\stackrel{p}{\longrightarrow}}

\global\long\def\dgoto{\stackrel{d}{\longrightarrow}}

\global\long\def\ll{\mathit{l}}

\global\long\def\logll{\mathcal{L}}

\global\long\def\bzero{\vt0}

\global\long\def\bone{\mathbf{1}}

\global\long\def\bff{\vt f}

\global\long\def\bx{\boldsymbol{x}}

\global\long\def\bX{\boldsymbol{X}}

\global\long\def\bW{\mathbf{W}}

\global\long\def\bH{\mathbf{H}}

\global\long\def\bL{\mathbf{L}}

\global\long\def\tbx{\tilde{\bx}}

\global\long\def\by{\boldsymbol{y}}

\global\long\def\bY{\boldsymbol{Y}}

\global\long\def\bz{\boldsymbol{z}}

\global\long\def\bZ{\boldsymbol{Z}}

\global\long\def\bu{\boldsymbol{u}}

\global\long\def\bU{\boldsymbol{U}}

\global\long\def\bv{\boldsymbol{v}}

\global\long\def\bV{\boldsymbol{V}}

\global\long\def\bw{\vt w}

\global\long\def\balpha{\gvt\alpha}

\global\long\def\bbeta{\gvt\beta}

\global\long\def\bmu{\gvt\mu}

\global\long\def\btheta{\boldsymbol{\theta}}

\global\long\def\blambda{\boldsymbol{\lambda}}

\global\long\def\realset{\mathbb{R}}

\global\long\def\realn{\real^{n}}

\global\long\def\natset{\integerset}

\global\long\def\interger{\integerset}

\global\long\def\integerset{\mathbb{Z}}

\global\long\def\natn{\natset^{n}}

\global\long\def\rational{\mathbb{Q}}

\global\long\def\realPlusn{\mathbb{R_{+}^{n}}}

\global\long\def\comp{\complexset}
 \global\long\def\complexset{\mathbb{C}}

\global\long\def\and{\cap}

\global\long\def\compn{\comp^{n}}

\global\long\def\comb#1#2{\left({#1\atop #2}\right) }
\selectlanguage{british}%

\begin{abstract}
In this paper, we develop a Bayesian optimization based hyperparameter
tuning framework inspired by statistical learning theory for classifiers.
We utilize two key facts from PAC learning theory; the generalization
bound will be higher for a small subset of data compared to the whole,
and the highest accuracy for a small subset of data can be achieved
with a simple model. We initially tune the hyperparameters on a small
subset of training data using Bayesian optimization. While tuning
the hyperparameters on the whole training data, we leverage the insights
from the learning theory to seek more complex models. We realize this
by using directional derivative signs strategically placed in the
hyperparameter search space to seek a more complex model than the
one obtained with small data. We demonstrate the performance of our
method on the tasks of tuning the hyperparameters of several machine
learning algorithms. 
\end{abstract}

\section{Introduction\label{sec:intro}}

\selectlanguage{english}%
Hyperparameter tuning is a challenging problem in machine learning.
Bayesian optimization has emerged as an efficient framework for hyperparameter
tuning, outperforming most conventional methods such as grid search
and random search \cite{Snoek2012,shahriari2016taking}. It offers
robust solutions for optimizing expensive black-box functions, using
a non-parametric Gaussian Process \cite{williams2006gaussian} as
a probabilistic measure to model the unknown function. A surrogate
utility function, known as acquisition function guides the search
for the next observation. The acquisition function continually trades-off
two key aspects: exploring regions where epistemic uncertainty about
the function is high and exploiting regions where the predictive mean
is high. The use of Bayesian optimization for efficient hyperparameter
tuning of complex models have been first proposed in \cite{Snoek2012}.
Despite the advances, hyperparameter tuning on large datasets remains
challenging.

Early attempt to make Bayesian optimization efficient includes multi-task
Bayesian optimization \cite{Swersky2013} to optimize hyperparameters
of the whole data in presence of an auxiliary problem of hyperparameter
optimization for a small subset of data. They proceed by estimating
task relationship, then finding best parameters for these tasks simultaneously
within the same framework. First, it is inefficient because both the
cheap and expensive processes are being discovered simultaneously,
making early estimate of their relationship poor, and thus delaying
convergence. Second, it is not straightforward to design a task-to-task
covariance function to leverage on how the hyperparameters are related
across these two tasks. 

An alternate approach has been attempted in \cite{pmlr-v54-klein17a}
where they assume that performance of a hyperparameter on a smaller
subset of data can be used to predict the performance of the same
hyperparameter on the whole data. This frees them from directly evaluating
on the larger dataset. Their assumption, however, is poorly grounded
on learning theory. The authors of \cite{pmlr-v70-kandasamy17a} developed
a similar strategy where a multi-fidelity optimization technique has
been used to approximate the expensive functions. Their work also
have not leveraged any task-specific knowledge from learning theory.
Recently, authors of \cite{li2016hyperband} have developed a multi-arm
bandit based strategy, called Hyperband algorithm. Hyperband uses
a strategy that allocates a finite resource (data samples, iteration)
budget to randomly sampled configurations, and choose the configurations
that are expected to perform well based on their performance on the
subset of data. Hyperband makes sure that only promising hyperparameters
proceed to the next round, and less promising configurations are discarded.
Their hypothesis, however, does not hold for hyperparameters that
directly control the model complexity, e.g. cost parameter C in support
vector machines (SVM). One instance of such a hyperparameter may work
well on a subset of data but the same may perform poorly on the larger
data. Time as resource may be useful when a gradient descent type
of algorithm can be used, but not for other kinds of optimization
routines as many do not guarantee smooth convergence. This restricts
the application of Hyperband to tuning hyperparameters that are functions
of data topology (learning rate, momentum, etc.), and when the model
is trained using gradient descent based algorithms. 

We seek our inspiration from PAC learning theory \foreignlanguage{british}{\cite{vapnik1999overview}}
and investigate an alternative approach for hyperparameter tuning
by utilizing key concepts about model complexity and generalization
performance with respect to dataset size:
\begin{itemize}
\item Vapnik--Chervonenkis inequality - The bound on the difference between
empirical error and the \emph{generalization error} is \emph{higher}
\emph{for smaller} datasets.
\item Vapnik--Chervonenkis dimension - For a small dataset, the \emph{smallest
generalization error} can be achieved with a \emph{lower complexity}
classifier.
\begin{figure}
\begin{centering}
\includegraphics[width=0.4\paperwidth]{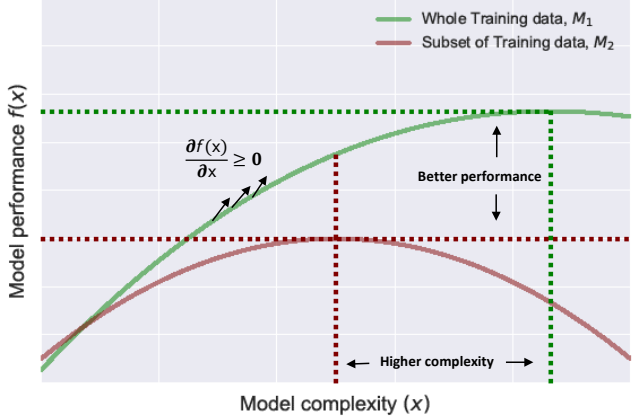}
\par\end{centering}
\centering{}\caption{Schematic representation of generalization performance vs model complexity.
We indicate the best performance of the two models and the corresponding
model complexities. \label{fig:hyper_per-1}}
\end{figure}
\end{itemize}
We illustrate our intuition in a schematic diagram of generalization
performance vs model complexity (Figure \ref{fig:hyper_per-1}). Two
models are built - one that uses the whole training data ($M_{1}$)
and the other only a subset of data ($M_{2}$). The idea is to transfer
some knowledge from $M_{2}$ to $M_{1}$. As expected, $M_{1}$ seeks
more complex models achieving higher performance - this performance
increases monotonically till it saturates. $M_{2}$ on the other hand
settles on a lower complex configuration and its generalization performance
degrades as it selects more complex models. $M_{2}$ exhibits poorer
generalization performance compared to $M_{1}$. Between the lower
and upper bounds of the performance in $M_{2}$, the performance of
$M_{1}$ rises strictly monotonically. We can use this extra knowledge
to tune the hyperparameters of $M_{1}$. We do this by introducing
signs of the derivatives \cite{riihimaki2010gaussian} (instead of
the actual derivative as these are unavailable) as additional observations
to the GP modeling $M_{1}$, and show that this results in a faster
hyperparameter tuning algorithm. Using this intuition we construct
a method termed \emph{HyperTune}.

We demonstrate the empirical performance of our method on the task
of tuning hyperparameters of several machine learning algorithms on
real-world datasets. We compare our results against the baselines
of Fabolas \cite{pmlr-v54-klein17a} and Hyeprband \cite{li2016hyperband}
and show that our algorithm performs better in almost all the tasks.\selectlanguage{british}%

\section{Key insights from PAC Generalization Bound by Vapnik and Chervonenkis\label{sec:bg-1}}

We note the main insights in \cite{vapnik1999overview} which relates
the generalization performance and complexity of the classification
models. VC dimension \cite{vapnik1999overview} of a hypothesis class
$\mathcal{H}\subseteq{0,1}^{\mathcal{X}}$ is defined as: 

\[
V_{\mathcal{H}}:=\max\left\{ n\mid S_{\mathcal{H}}(n)=2^{n}\right\} 
\]
where $S_{\mathcal{H}}(n)$ is the shatter coefficient \cite{vapnik1999overview}
and $n$ denotes the number of data points, and is defined as,
\[
S_{\mathcal{H}}(n)=:\max_{\mathbf{x}_{1},...,\mathbf{x}_{n}\in\mathcal{X}}N_{\mathcal{H}}(\mathbf{x}_{1},...,\mathbf{x}_{n})
\]
where $\mathbf{x}_{1},...,\mathbf{x}_{n}$ denote data points, and
$N_{\mathcal{H}}(\mathbf{x}_{1},...,\mathbf{x}_{n}):=|\{(h(\mathbf{x}_{1}),...,h(\mathbf{x}_{n})):h\in\mathcal{H}\}|$.
We can relate VC dimension and shatter coefficient using a generalization
bound. We recall the theorem and corollary in \cite{vapnik1999overview}
as:
\begin{thm}
\cite{vapnik1999overview} For binary classification and the 0/1 loss
function, we have the following generalization bounds,
\[
\Pr\left(sup_{h\in\mathcal{H}}\mid\hat{R_{n}}(h)-R(h)\mid>\varepsilon\right)\leq8S_{\mathcal{H}}\left(n\right)e^{\frac{-ne^{2}}{32}}
\]
where the probability is with respect to the draw of the training
data. Here $\hat{R_{n}}(h)$ denotes the empirical risk and $R(h)$
denotes the expected generalization risk for a hypothesis $h$.
\end{thm}
\begin{cor}
\cite{vapnik1999overview} If \textup{$\hat{R_{n}}(h)$} is an empirical
risk over $\mathcal{H}$, then with probability greater than $1-\delta$,
the generalization bound is given as{\small{},
\[
\hat{R_{n}}(h)\leq R(h)+\sqrt{\dfrac{128\left[V_{\mathcal{H}}\log(n+1)\right]+\log(\frac{8}{\delta})}{n}}
\]
}{\small\par}
\end{cor}
From the above Theorem and Corollary, we can conclude that if $n$
is small, then $V_{\mathcal{H}}$ should also be small. For a fixed
VC dimension $V_{\mathcal{H}}$, the generalization bound is higher
when $n$ is small. It also implies that the lowest generalization
risk (highest accuracy) for a model that is trained on a larger data
will occur at a more complex model than the one with a smaller training
set.

\section{Bayesian Optimization\label{sec:bg}}

\selectlanguage{english}%
Bayesian optimization is an elegant framework for sequential optimization
of unknown objective functions and can be summarized as, $\mathbf{x}^{*}=\arg\max_{\mathbf{x\in\mathcal{X}}}f(\mathbf{x})$,
where $\mathcal{X}\subseteq\mathbb{R}^{D}$. The function evaluations
can be noisy with $y=f(\mathbf{x})+\epsilon$ where $\epsilon\sim\mathcal{N}(0,\sigma^{2})$.
Gaussian process (GP) \cite{williams2006gaussian} is a popular choice
for specifying prior over smoothly varying functions. Let $\mathbf{x}$
be a $D$ dimensional observation and $\mathbf{X}$ be a matrix with
$p$ such observations. Let $\mathbf{y}$ be the corresponding function
values. Without any loss of generality, the prior mean function can
be assumed to be a zero function making the Gaussian process fully
defined by the covariance function as: $p\left(\text{\ensuremath{\mathbf{f}}}\mid\mathbf{X}\right)=\mathcal{N}(\mathbf{f}\mid0,\mathbf{K})$
where $\mathbf{f}$ is the corresponding latent variables and $\mathbf{K}$
is the kernel matrix with $\mathbf{K_{\text{i,j}}=k(x_{\text{i}},x_{\text{j}})}$.
We choose the popular squared exponential kernel as our choice of
the covariance function. It is defined as, $k(\mathbf{\mathbf{x},\mathbf{x'}})=exp(-\frac{1}{2\theta}||\mathbf{x}-\mathbf{x'}||^{2})$
where $\theta$ is the length-scale parameter of the kernel. Using
the properties of Gaussian Process, partial derivatives of a GP is
still Gaussian since differentiation is a linear process \cite{rasmussen2003gaussian,solak2003derivative}.
This allows us to include the derivative of the unknown objective
function using the following criteria for covariance functions as:

{\small{}
\begin{equation}
k\left[\frac{\partial f^{(i)}}{\partial x_{d}^{(i)}},f^{(j)}\right]=\frac{\partial}{\partial x_{d}^{(i)}}k\left[f^{(i)},f^{(j)}\right]
\end{equation}
}{\small\par}

{\small{}
\begin{equation}
k\left[\frac{\partial f^{(i)}}{\partial x_{d}^{(i)}},\frac{\partial f^{(j)}}{\partial x_{g}^{(j)}}\right]=\frac{\partial^{2}}{\partial x_{d}^{(i)}\partial x_{g}^{(j)}}k\left[f^{(i)},f^{(j)}\right]
\end{equation}
}where $d,g\in[1,...,D]$. Bayesian optimization now uses an acquisition
function that guides the search for the optimum of the underlying
objective function. In this paper, we use the expected improvement
(EI) \cite{mockus1978application} acquisition for its usefulness
and simplicity. However, it has to be noted that our method can work
with any acquisition function. EI is defined as: {\small{}$\alpha(\mathbf{x}_{p+1})=(\mu(\mathbf{x}_{p+1})-f(\mathbf{x}^{+}))\Phi(z)+\sigma(\mathbf{x}_{p+1})\phi(z)$
where $z=(\mu(\mathbf{x}_{p+1})-f(\mathbf{x}^{+}))/\sigma(\mathbf{x}_{p+1})$,
$\Phi(.)$ and $\phi(.)$ }are the CDF and PDF of a standard normal
distribution. \selectlanguage{british}%

\section{HyperTune - A Learning Theory based Hyperparameter Tuning Framework}

Consider a hyperparameter tuning task wherein our objective is to
maximize validation set accuracy as,
\[
\mathbf{x^{*}}=\arg\max_{\mathbf{x\in\mathcal{X}}}f(\mathbf{x})
\]
where $\mathcal{X}\subseteq\mathbb{R}^{D}$ and $f(\mathbf{x})$ is
performance of the model on validation dataset for a set of hyperparameters
$\mathbf{x}$. Let the search bound for different hyperparameters
be $[\mathbf{l},\mathbf{u}]$ - $\mathbf{l}$ and $\mathbf{u}$ are
$D$ dimensional vectors that denote the lower and the upper bounds,
respectively. 

While our objective is to optimize hyperparameters on the whole training
data, we start by identifying the optimal hyperparameters on a small
subset of the training data using Bayesian optimization. These optimal
hyperparameters can be noisy as they are built on a small portion
of data. We thus pick a robust estimate of the hyperparameter by repeating
Bayesian optimization on a number of different smaller subsets and
then average the optimal hyperparameters. The performance of the model\foreignlanguage{english}{
trained on small data (randomly created) may exhibit spurious pikes
that peak either at extremely high complex region or at a very low
complex region. In the former case, we may miss the optimal for the
larger data whilst the latter turns our algorithm unnecessarily inefficient}.
Through averaging, we smooth out the spurious behaviours and identify
the right complexity for the small data, which is then used to provide
directional derivative to effectively have a tighter bound for the
larger data. We denote this best hyperparameter from the smaller subset
as $\bar{\mathbf{x}}_{s}^{*}$.

\begin{algorithm}[t] 
\caption{Sampling Directional Derivative Observations.}
\label{alg:sampling} 
\begin{algorithmic}[1]   
\State Bounds on Search space : $[\mathbf{l},\mathbf{u}]$
\State Number of virtual observations : $N$
\For {$b=1,2,..B$} 
\State $\mathbf{x}_{s,b}^{*}$=$\arg\max_{\mathbf{x\in\mathcal{X}}}f_{s,b}(\mathbf{x})$ \Comment{Conduct Bayesian optimization on small subsets of data}
\EndFor 
\State $\bar{\mathbf{x}}_{s}^{*}=\ensuremath{\frac{1}{B}}\sum_{b=1}^{B}\mathbf{x}_{s,b}^{*}$ %\Comment{Selecting an optimal hyperparameter}
\State $\mathbf{X}_{m}\sim\mathcal{U}(L,\bar{\mathbf{x}}_{s}^{*},N)$ %\Comment{Uniform sampling of $N$ directional derivative obserations}
\For {$d=1,2,..D$}
	\If{$\dfrac{\partial f(\mathbf{x})}{\partial x_{d}}\geq0$}      
		\State $m_{d}=1$
	\ElsIf{$\dfrac{\partial f(\mathbf{x})}{\partial x_{d}}\leq0$} 
		\State $m_{d}=-1$ 
	\Else
		\State $m_{d}=null$ 	
	\EndIf
\EndFor
\end{algorithmic}%  
\end{algorithm}

While tuning hyperparameters on the whole dataset, we note that for
certain hyperparameters which directly controls the model complexity,
the generalization performance would be monotonically changing in
the bound $[\mathbf{l},\bar{\mathbf{x}}_{s}^{*}]$. If a particular
hyperparameter increases the complexity, then the model performance
also increases for a larger dataset wherein it decreases otherwise.
For example, the model complexity of SVM increases as the cost parameter
$C$ increases.\foreignlanguage{english}{ The model complexity of
elastic net \cite{zou2005regularization} decreases with an increase
in the hyperparameter $\alpha$ that determines the magnitude of the
penalty term.} We exploit such trends by sampling $N$ directional
derivative observations $\mathbf{X}_{m}$ uniformly within the range
$[\mathbf{l},\bar{\mathbf{x}}_{s}^{*}]$\@. Using the expert knowledge,
we assign a sign that encodes our belief about how the trend will
be changing for a model that uses the whole data. Let $\mathbf{m}$
be a vector of directional derivative signs that takes only $1$ or
$-1$ indicating whether the function is either monotonically increasing
or decreasing. we assign $m_{d}=1$ when $\ensuremath{\dfrac{\partial f(\mathbf{x})}{\partial x_{d}}\geq0}$
and $m_{d}=-1$ when $\ensuremath{\dfrac{\partial f(\mathbf{x})}{\partial x_{d}}\leq0}$.
We consider a hyperparameter as neutral if it does not contribute
towards model complexity, in which case we do not encode any directional
sign (e.g. learning rate for neural networks). We further detail the
procedure for sampling the directional derivatives in Algorithm \ref{alg:sampling}.

\begin{algorithm}[t]   
\caption{\textit{HyperTune}.}
\label{alg:NBO-1-1} 
\begin{algorithmic}[1]
\State Input: Initial observations $\{\mathbf{X},\mathbf{y}\}$
\State Directional Derivative Observations $\{\mathbf{X}_{m},{\mathbf{m}}\}$ \Comment{Using Algorithm \ref{alg:sampling}}
%\State $\mathbf{X}=\{\mathbf{X},\mathbf{X}_{m}\}$, $\boldsymbol{y}=\{\boldsymbol{y},\mathbf{m}\}$
\State Combined observations $O\equiv\{\mathbf{X},\mathbf{X}_{m},\mathbf{y},\mathbf{m}\}$
\For {$t=1,2,..T$} 
\State $\mathbf{x}_{t}=\arg\max_{\mathbf{x\in\mathcal{X}}}\alpha(\mathbf{x}|O)$
\State Evaluate $f(.)$ at $\mathbf{x}_{t}$
\State $O\equiv{O\cup(\mathbf{x}_{t},\mathbf{y}_{t})}$ 
\EndFor  
\State Output: $\{\mathbf{x}_{t},\mathbf{y}_{t}\}_{t=1}^{T}$
\end{algorithmic}%  
\end{algorithm} 

Let $\mathbf{X}$ be an initial set of  hyperparameters and $\mathbf{y}$
be the corresponding model performances on the whole data. Let $\mathbf{f}$
be the latent function values and $\mathbf{f}^{'}$ be the corresponding
derivative values. The joint prior for latent values and derivatives
is also Gaussian as, 
\[
p\left(\mathbf{f},\mathbf{f}'\mid\mathbf{X},\mathbf{X}_{m}\right)=\mathcal{N}(\mathbf{f}_{joint}\mid0,\mathbf{K}_{joint})
\]
where $\mathbf{f}_{joint}$ and $\mathbf{K}_{joint}$ are given as,

\begin{equation}
\mathbf{f}_{joint}=\left[\begin{array}{c}
\mathbf{f}\\
\mathbf{f}^{'}
\end{array}\right],\mathbf{K}_{joint}=\left[\begin{array}{cc}
\mathbf{K_{\mathbf{f},\mathbf{f}}} & \mathbf{K_{\text{\ensuremath{\mathbf{f}}},\mathbf{f}^{'}}}\\
\mathbf{K_{\mathbf{f}^{'},\text{\text{\ensuremath{\mathbf{f}}}}}} & \mathbf{K_{\mathbf{f}^{'},\mathbf{f}^{'}}}
\end{array}\right]\label{eq:eq_toref_transf}
\end{equation}
 For a new point $\mathbf{x}_{p+1}$, the predictive posterior distribution
of the Gaussian process can now be expressed as,

\selectlanguage{english}%
{\footnotesize{}
\begin{align*}
p(f\mid\mathbf{x}_{p+1},\mathbf{X},\mathbf{y},\mathbf{X}_{m},\mathbf{m}))=\int p\left(f\mid\mathbf{x}_{p+1},\mathbf{X},\mathbf{f},\mathbf{X}_{m},\mathbf{f}^{'}\right)\\
\times p\left(\mathbf{f},\mathbf{f}'\mid\mathbf{y},\mathbf{m}\right)d\mathbf{f}d\mathbf{f}^{'}
\end{align*}
}\foreignlanguage{british}{Using the property of GP, the joint posterior
distribution can now be defined as,}

\selectlanguage{british}%
{\small{}
\begin{equation}
p\left(\mathbf{f},\text{\ensuremath{\mathbf{f}}}'\mid\mathbf{y},\mathbf{m}\right)=\frac{1}{Z}p\left(\mathbf{f},\mathbf{f}'\mid\mathbf{X},\mathbf{X}_{m}\right)p\left(\mathbf{y}\mid\mathbf{f}\right)p\left(\mathbf{m}\mid\mathbf{f}^{'}\right)\label{eq:partial_derive_pred}
\end{equation}
}Here $Z$ is a normalization term and is given as,

\[
Z=\int p\left(\mathbf{f},\mathbf{f}^{'}\mid\mathbf{X},\mathbf{X}_{m}\right)p\left(\mathbf{y}\mid\mathbf{f}\right)p\left(\mathbf{m}\mid\mathbf{f}^{'}\right)d\mathbf{f}d\mathbf{f}^{'}
\]
We now express the probability of directional derivative signs $p\left(m\mid\mathbf{f}^{'}\right)$
using a Probit observation model \cite{riihimaki2010gaussian} as,

\begin{equation}
p\left(\mathbf{m}\mid\mathbf{f}^{'}\right)=\prod_{i=1}^{M}\Phi\left(\dfrac{m_{i}\partial f^{(i)}}{\partial x_{d}^{(i)}}\frac{1}{v}\right)\label{eq:probit}
\end{equation}
{\small{}where $\Phi\left(z\right)=\intop_{-\infty}^{z}\mathcal{N}(t\mid0,1)dt$},
and parameter $v$ controls the steepness of each step and thereby
the strictness of information about the partial derivatives. We use
$v=10^{-6}$ as suggested by \cite{riihimaki2010gaussian}. The expression
for joint predictive distribution in Equation (\ref{eq:partial_derive_pred})
becomes intractable due to the likelihood for the derivative observations
as given in Equation (\ref{eq:probit}). We use expectation propagation
\cite{riihimaki2010gaussian} algorithm to approximate the expression
in Equation (\ref{eq:partial_derive_pred}). We present \textit{HyperTune}
in Algorithm \ref{alg:NBO-1-1}.

\section{Experiments}

\selectlanguage{english}%
We evaluate the performance of \emph{Hypertune} on the tasks of tuning
the hyperparameters of five machine learning algorithms - elastic
net, SVM with rbf kernel, SVM with kernel approximation, multi-layer
perceptron (MLP), and convolutional neural network (CNN). We compare
our method with baselines Fabolas \cite{pmlr-v54-klein17a} and an
EI based generic Bayesian optimization algorithm on all the tasks.
\foreignlanguage{british}{Hyperband \cite{li2016hyperband} is compared
only with the last three tasks as these algorithms can be trained
using gradient descent-based algorithms that have smooth convergence.}

\subsection{Experimental Setting}

We use the implementation for Fabolas \cite{pmlr-v54-klein17a} and
Hyperband \cite{li2016hyperband} from their open source package,
RoBo\footnote{https://github.com/automl/RoBO}. We implement the
hyperparameter tuning tasks using an open source benchmark repository,
called HPOlib\footnote{https://github.com/automl/HPOlib2}.  A similar
experimental setup as in Fabolas is used. A validation dataset is
used for hyperparameter tuning, and a heldout set is used to evaluate
generalization performance. In all experiments, we visualize the convergence
using plots and report the performance on the held-out data using
a table. Fabolas uses subsets of data in the optimization loop, whereas
our method and a standard Bayesian optimization use full training
data. To obtain a fair convergence plot, we have only kept \emph{Hypertune}
and a generic Bayesian optimization algorithm in the plots. However,
we report the best performance returned by Fabolas after rebuilding
the model using the best hyperparameter setting from the optimization
loop on the full training data and evaluating it on the heldout data.

For each experiment, a small percentage of data is used to obtain
directional derivative information for \emph{HyperTune}. The hyperparameter
tuning on the smaller dataset is performed 5 times, and their running
time recorded. Although they can be computed in parallel, we consider
them to be computed sequentially here. The percentage of data for
the smaller dataset experiment is varied depending on the dataset.
Next, we run \emph{HyperTune} until convergence. The running time
for \emph{HyperTune} includes the time to converge to the best validation
set performance (not considering if the improvement is within the
$2\sigma$), and the time for smaller dataset experiments. This total
time is the budget for the other baselines and we compute the resultant
test errors using the best hyperparameter from all the methods on
a heldout dataset.

We use four benchmark real-world classification datasets from OpenML
\cite{vanschoren2014openml} - letter recognition with {\small{}20000}
instances and 17 features, MNIST digit recognition with 70000 instances
and 785 features, adult income prediction with {\small{}48842} instances
and 15 features, and vehicle registration with {\small{}98528} instances
and 101 features.
\begin{figure*}
\begin{centering}
\subfloat[\foreignlanguage{british}{}]{\centering{}\includegraphics[width=0.25\textwidth]{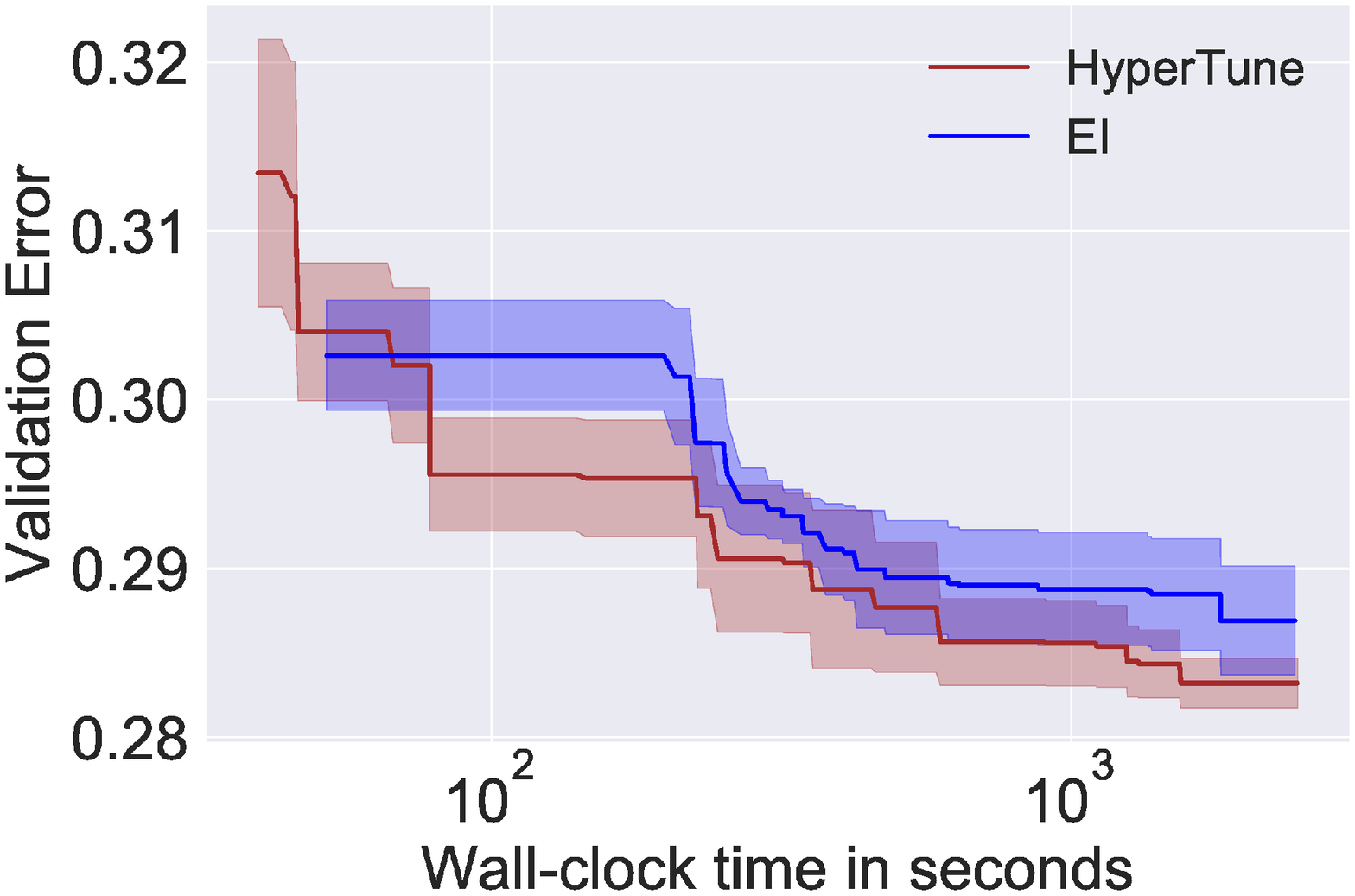}}\subfloat[\foreignlanguage{british}{}]{\centering{}\includegraphics[width=0.25\textwidth]{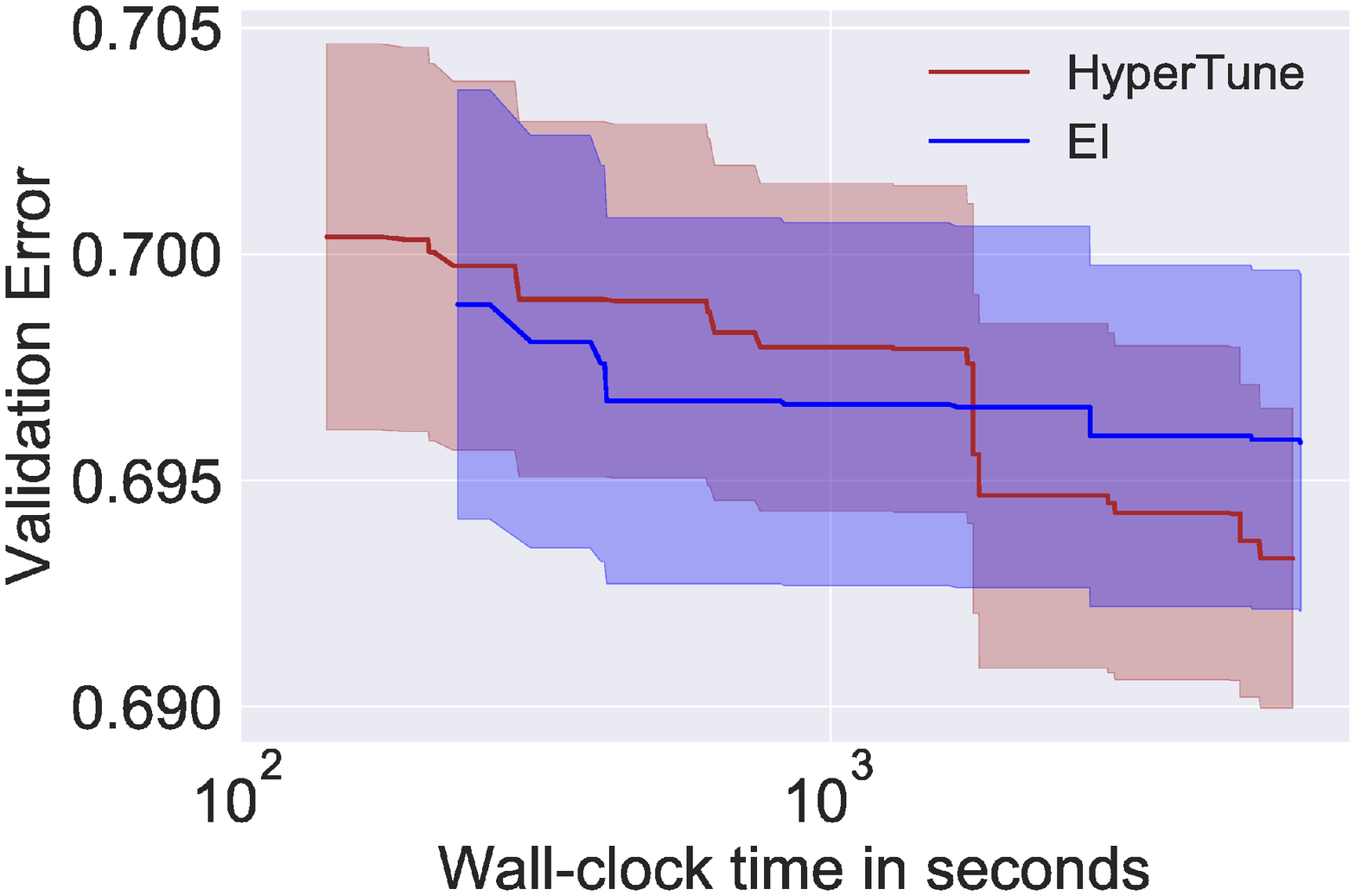}}\subfloat[\foreignlanguage{british}{}]{\centering{}\includegraphics[width=0.25\textwidth]{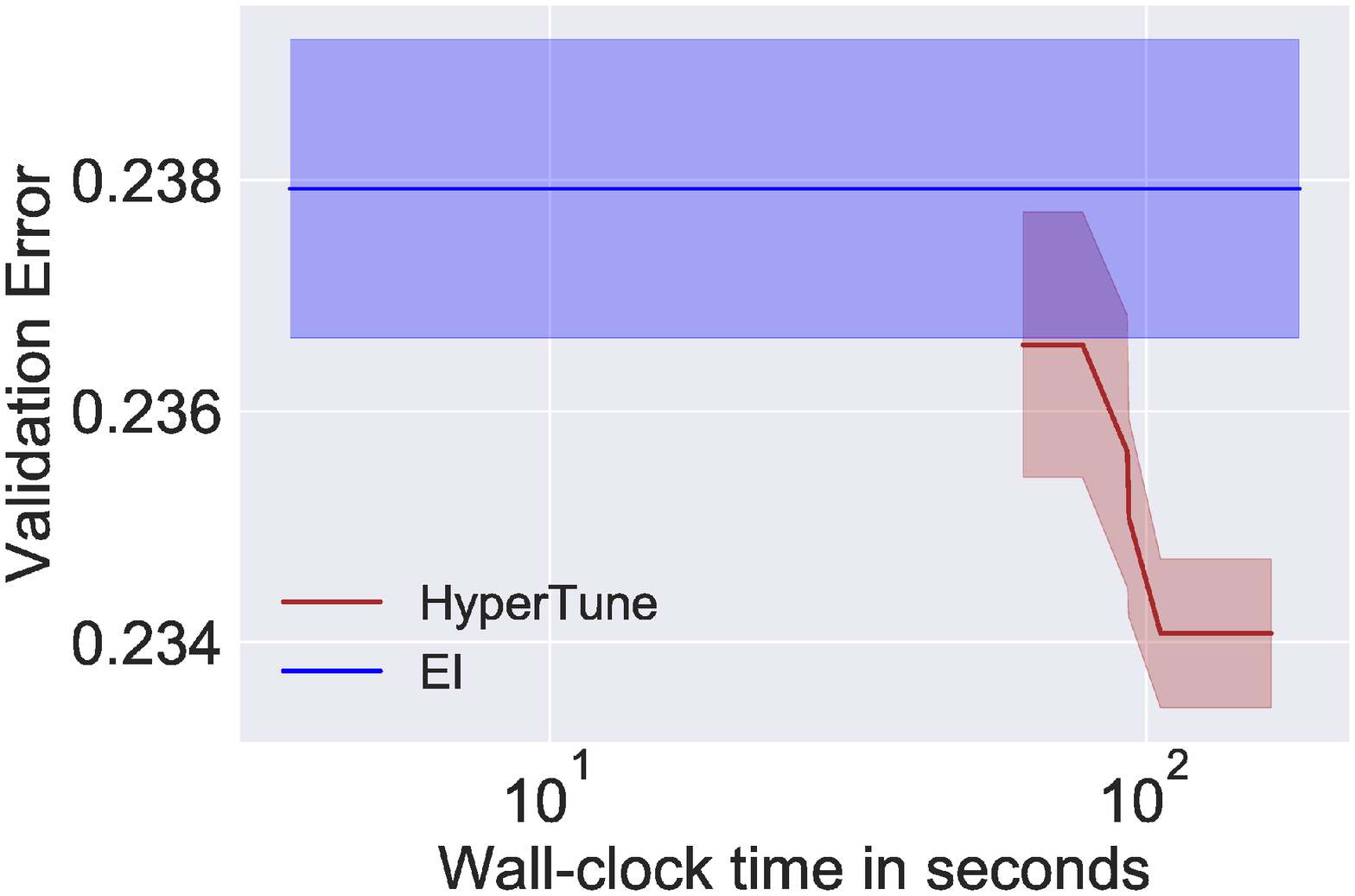}}\subfloat[\foreignlanguage{british}{}]{\centering{}\includegraphics[width=0.25\textwidth]{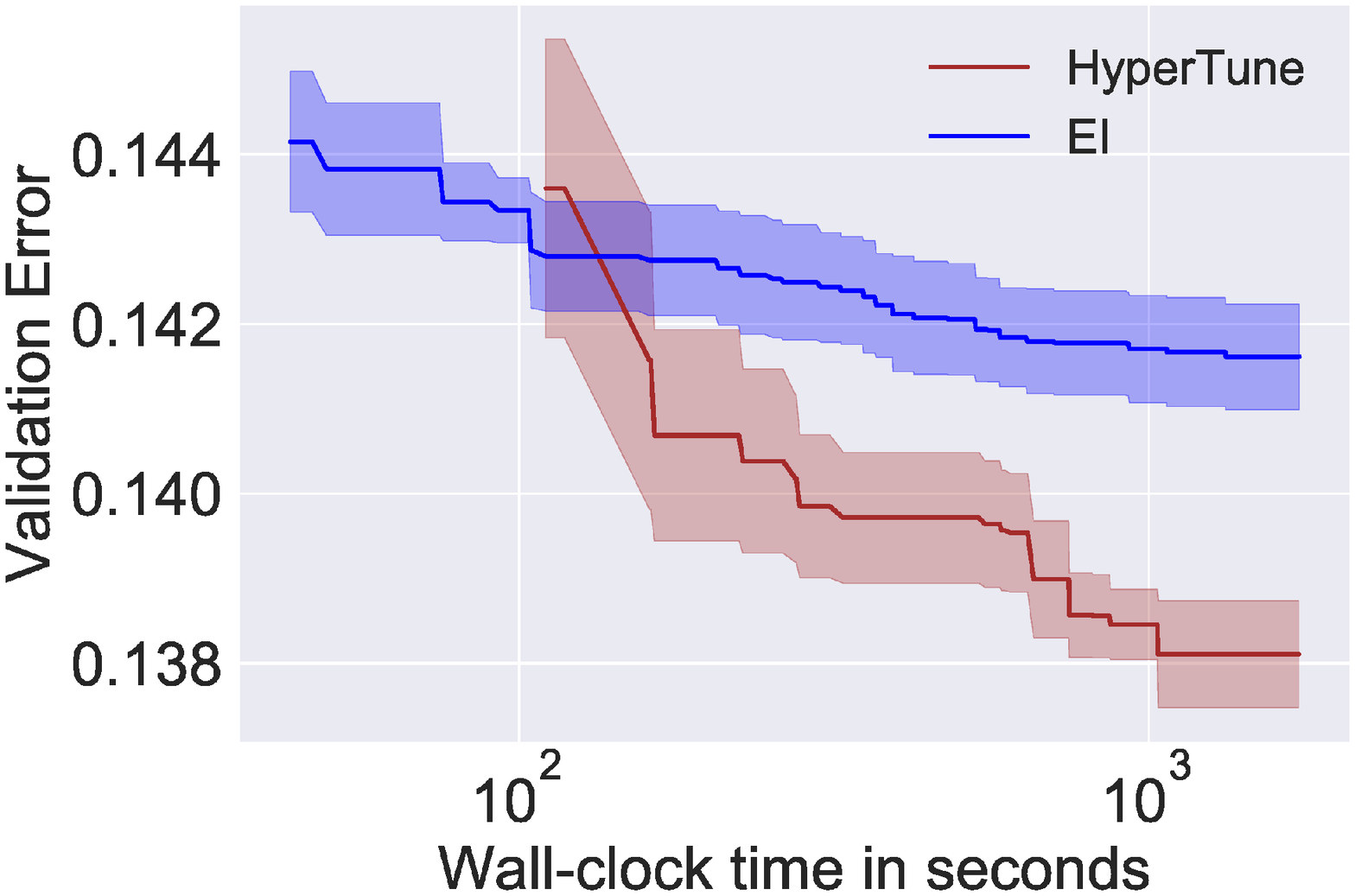}}
\par\end{centering}
\centering{}\caption{EI and \emph{HyperTune} on validation dataset for Elastic net: (a)
Letter, (b) MNIST, (c) Adult, and (d) Vehicle datasets. \label{fig:elnet_valid}}
\end{figure*}
\begin{table*}
\centering{}%
\begin{tabular}{|c|c|c|c|c|}
\hline 
\multirow{2}{*}{Baselines} & \multicolumn{4}{c|}{Average of the test error $\pm$ standard error}\tabularnewline
\cline{2-5} 
 & Letter & MNIST & Adult & Vehicle\tabularnewline
\hline 
Fabolas & {\small{}0.32$\pm$0.005} & {\small{}0.12$\pm$0.012} & {\small{}0.45$\pm$0.112} & {\small{}0.15$\pm$0.0005}\tabularnewline
\hline 
\textit{HyperTune} & \textbf{\small{}0.3$\pm$0.000} & \textbf{\small{}0.11$\pm$0.007} & \textbf{\small{}0.24$\pm$0.000} & \textbf{\small{}0.14$\pm$0.0004}\tabularnewline
\hline 
\end{tabular}\caption{Generalization performance on the heldout dataset for Elastic net.\label{tab:Elnet_test}}
\end{table*}
\begin{figure*}
\centering{}\subfloat[\foreignlanguage{british}{}]{\centering{}\includegraphics[width=0.25\textwidth]{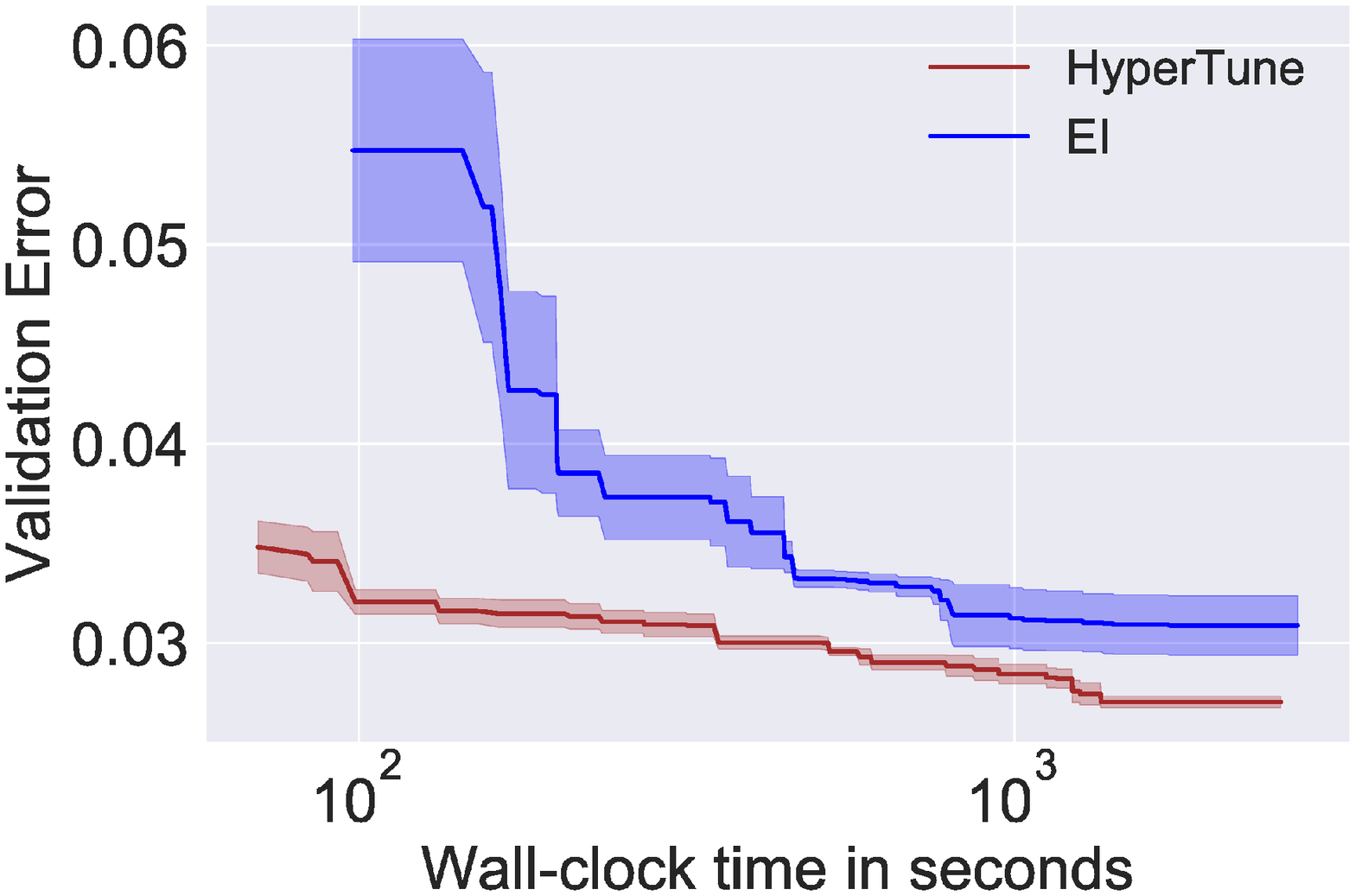}}\subfloat[\foreignlanguage{british}{}]{\centering{}\includegraphics[width=0.25\textwidth]{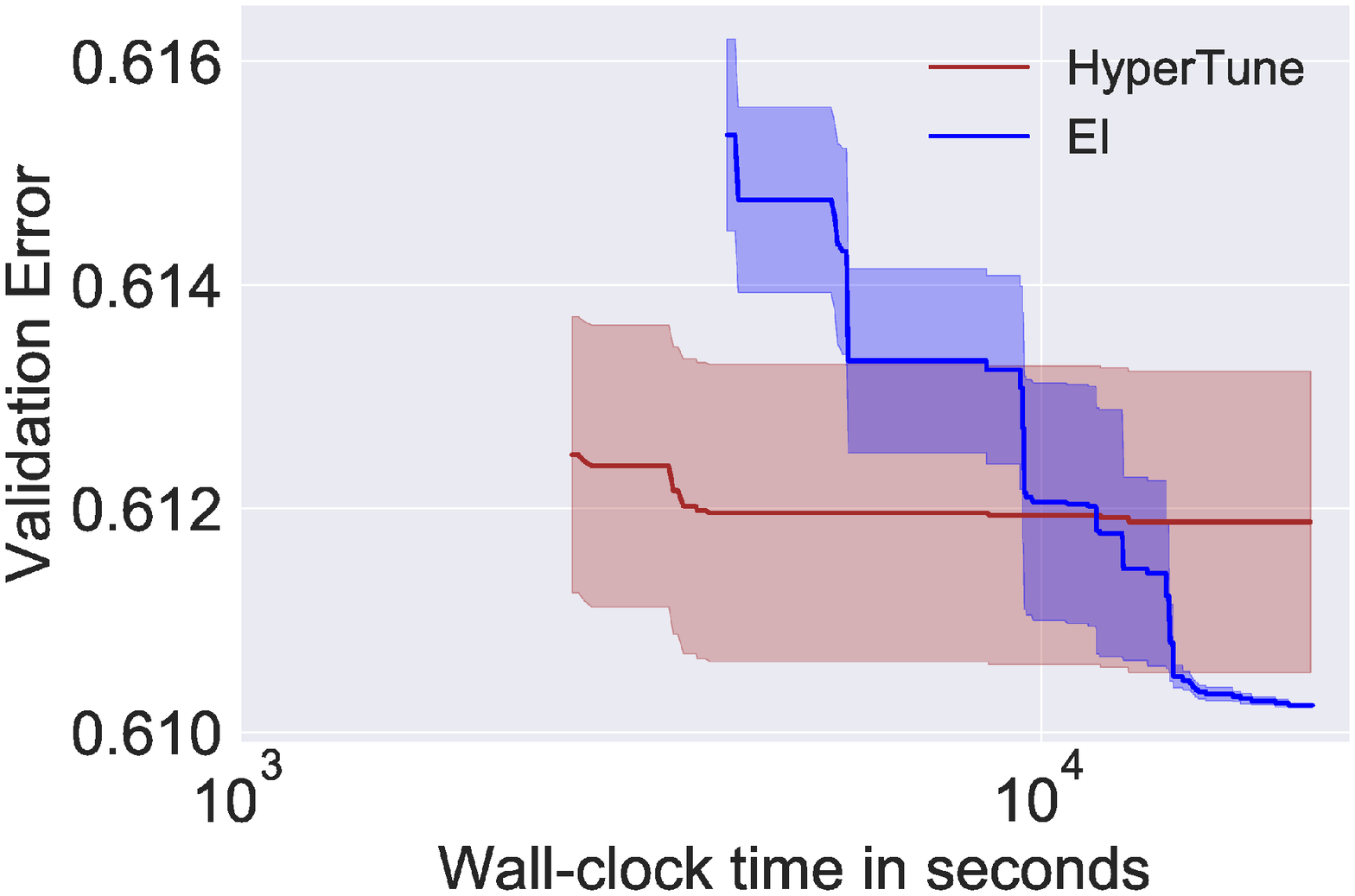}}\subfloat[\foreignlanguage{british}{}]{\centering{}\includegraphics[width=0.25\textwidth]{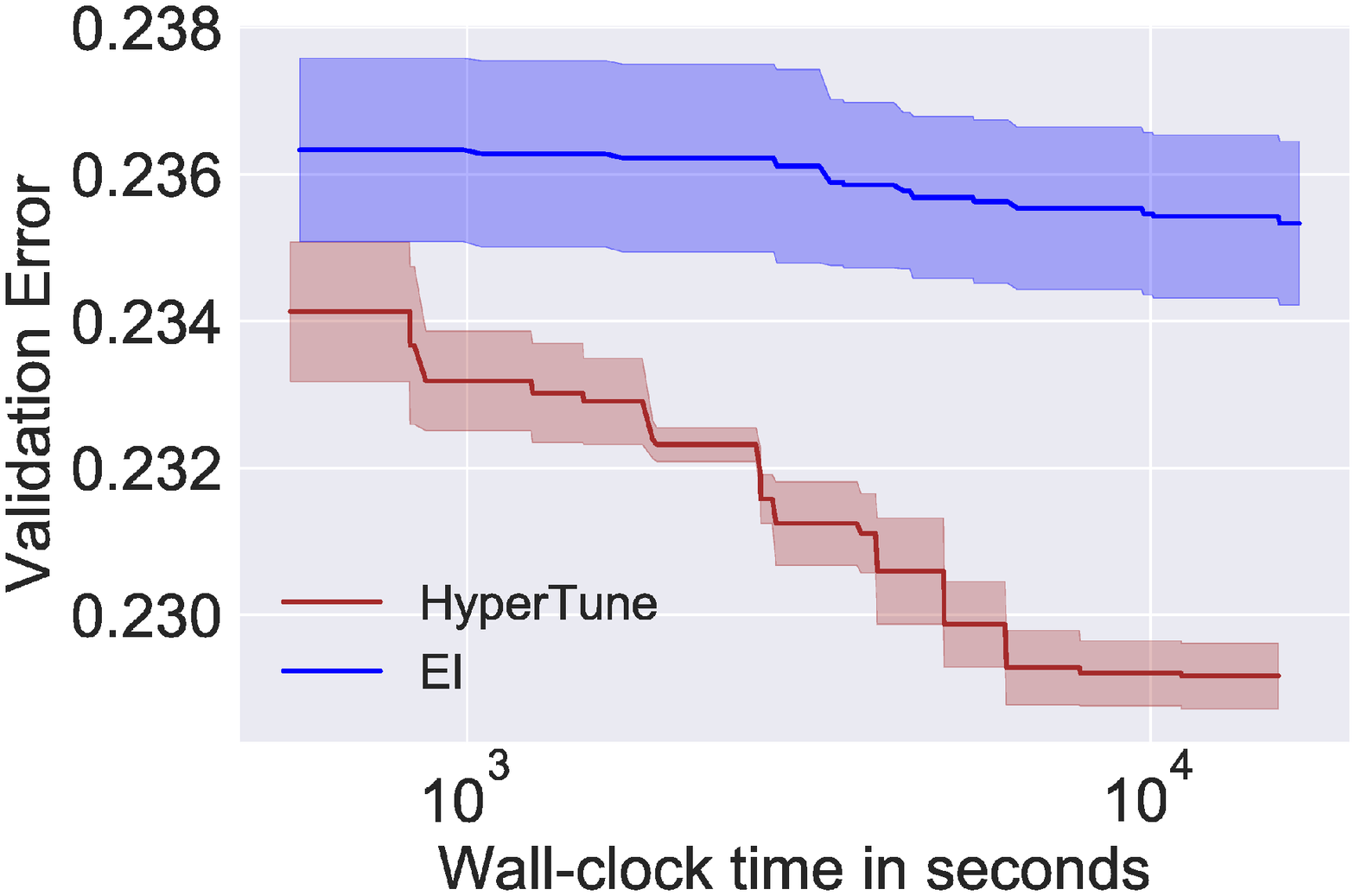}}\subfloat[\foreignlanguage{british}{}]{\centering{}\includegraphics[width=0.25\textwidth]{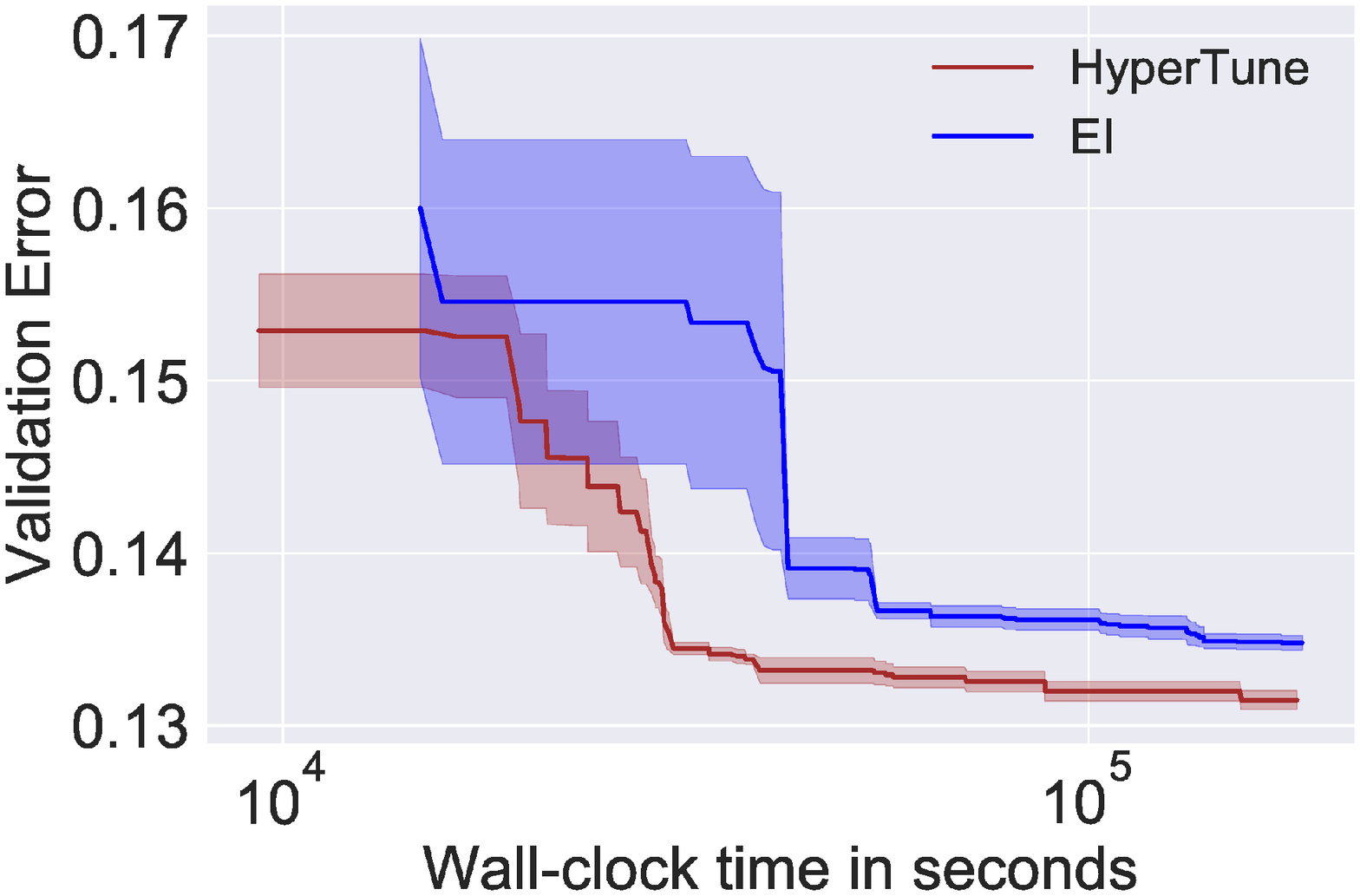}}\caption{EI and \textit{HyperTune} on validation dataset for SVM with RBF kernel:
a) Letter, (b) MNIST, (c) Adult, and (d) Vehicle datasets.\label{fig:svm_VALID} }
\end{figure*}
\begin{table*}
\centering{}%
\begin{tabular}{|c|c|c|c|c|}
\hline 
\multirow{2}{*}{Baselines} & \multicolumn{4}{c|}{Average of the test error $\pm$ standard error}\tabularnewline
\cline{2-5} 
 & Letter & MNIST & Adult & Vehicle\tabularnewline
\hline 
Fabolas & {\small{}0.54$\pm$0.180} & {\small{}0.016$\pm$0.0005} & \textbf{\small{}0.24$\pm$0.0005} & \textbf{\small{}0.14$\pm$0.002}\tabularnewline
\hline 
\textit{HyperTune} & \textbf{\small{}0.04$\pm$0.000} & \textbf{\small{}0.014$\pm$0.0002} & \textbf{\small{}0.24$\pm$0.0001} & \textbf{\small{}0.14$\pm$0.000}\tabularnewline
\hline 
\end{tabular}\caption{Generalization performance on heldout dataset for SVM with RBF kernel.\label{tab:svm_test}}
\end{table*}
\begin{figure*}
\centering{}\subfloat[\foreignlanguage{british}{}]{\centering{}\includegraphics[width=0.25\textwidth]{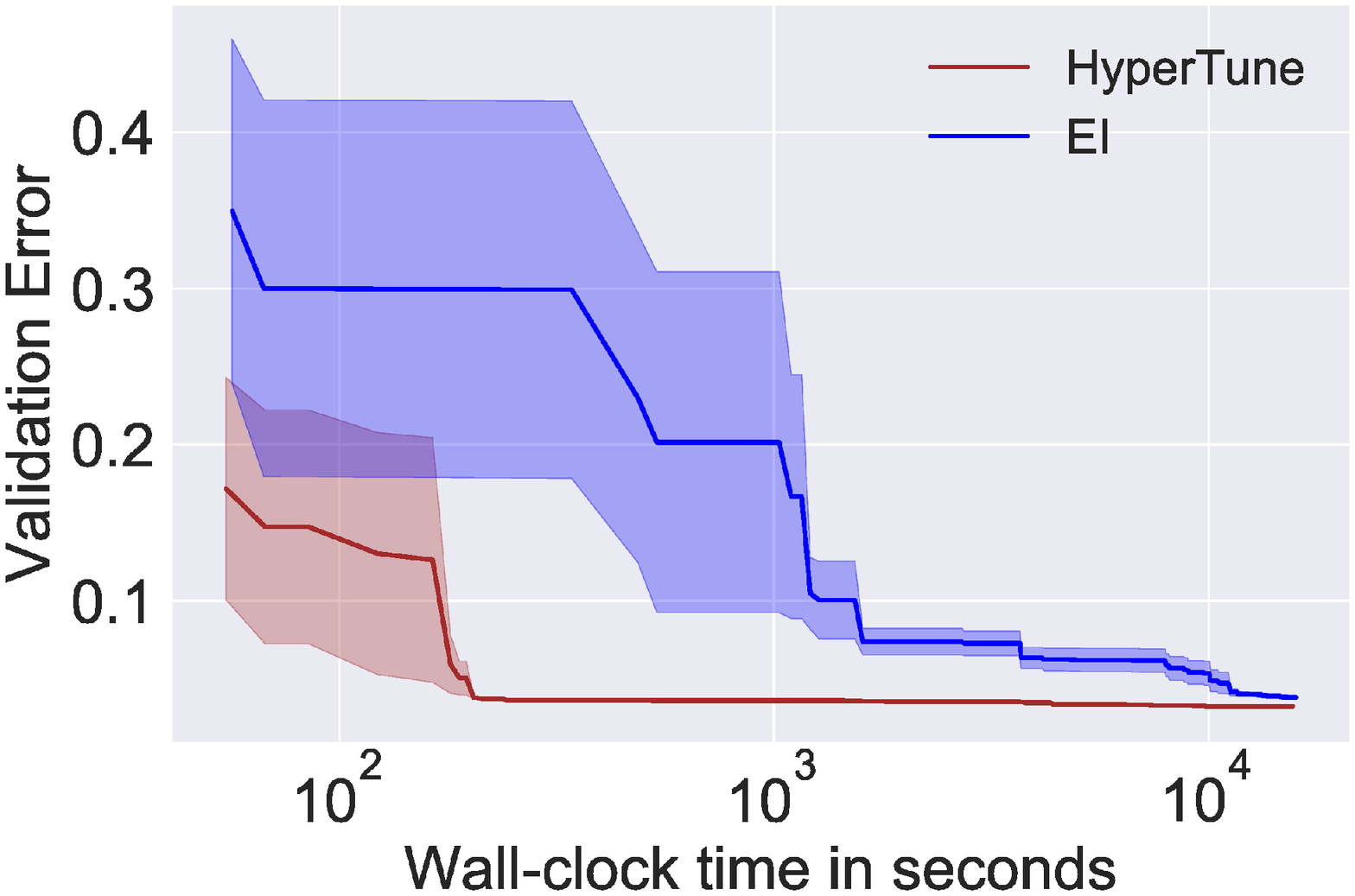}}\subfloat[\foreignlanguage{british}{}]{\centering{}\includegraphics[width=0.25\textwidth]{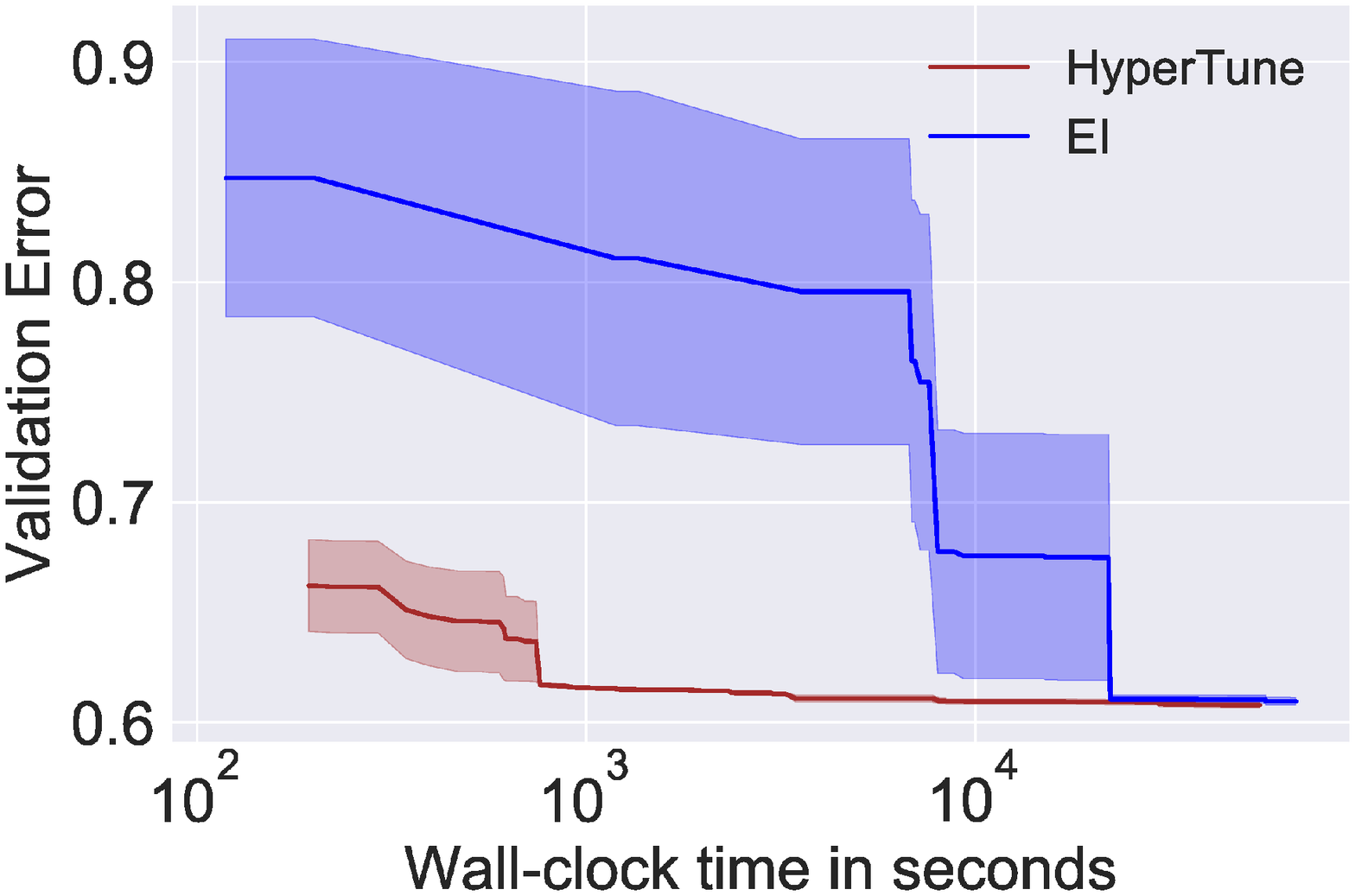}}\subfloat[\foreignlanguage{british}{}]{\centering{}\includegraphics[width=0.25\textwidth]{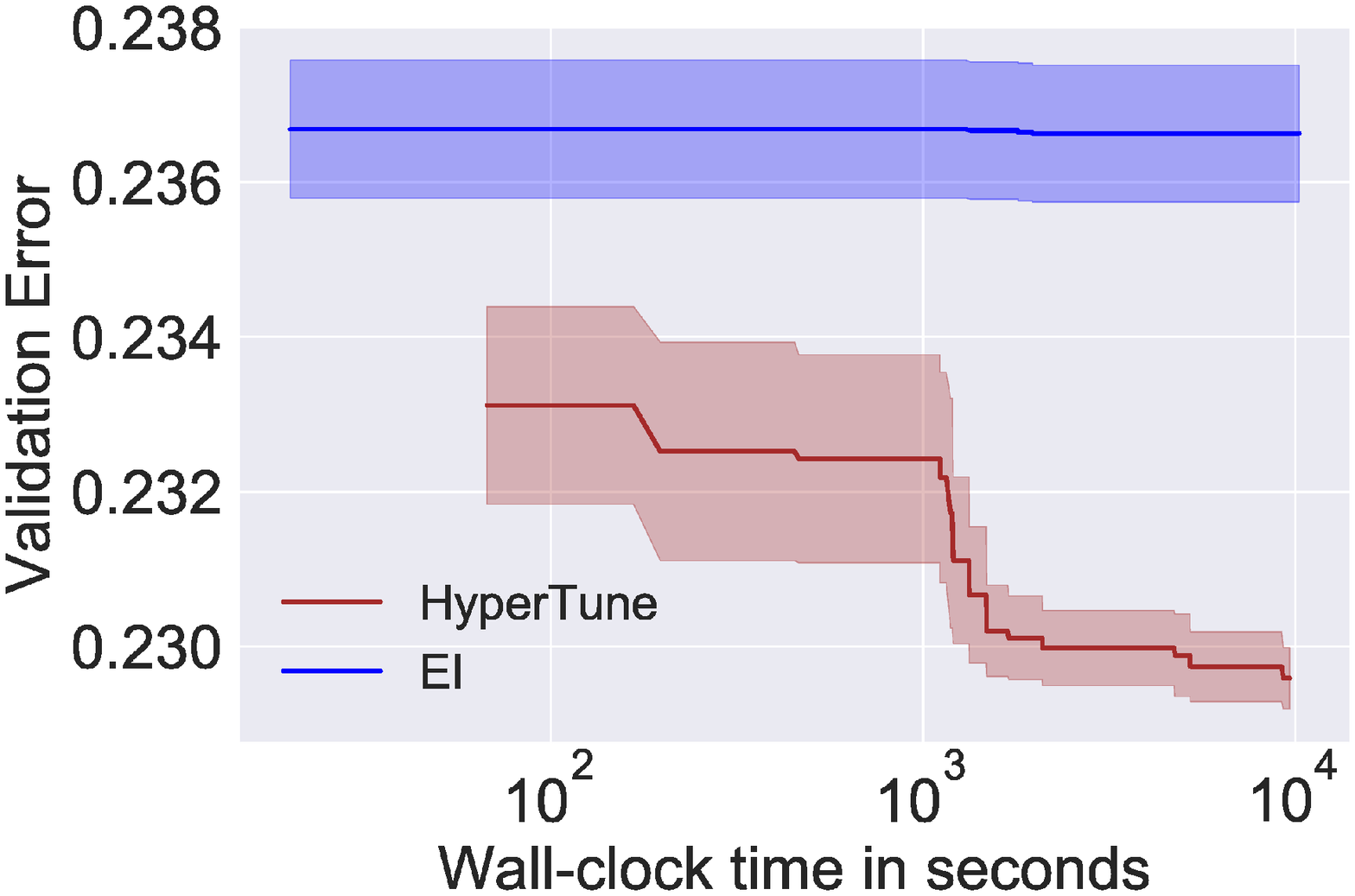}}\subfloat[\foreignlanguage{british}{}]{\centering{}\includegraphics[width=0.25\textwidth]{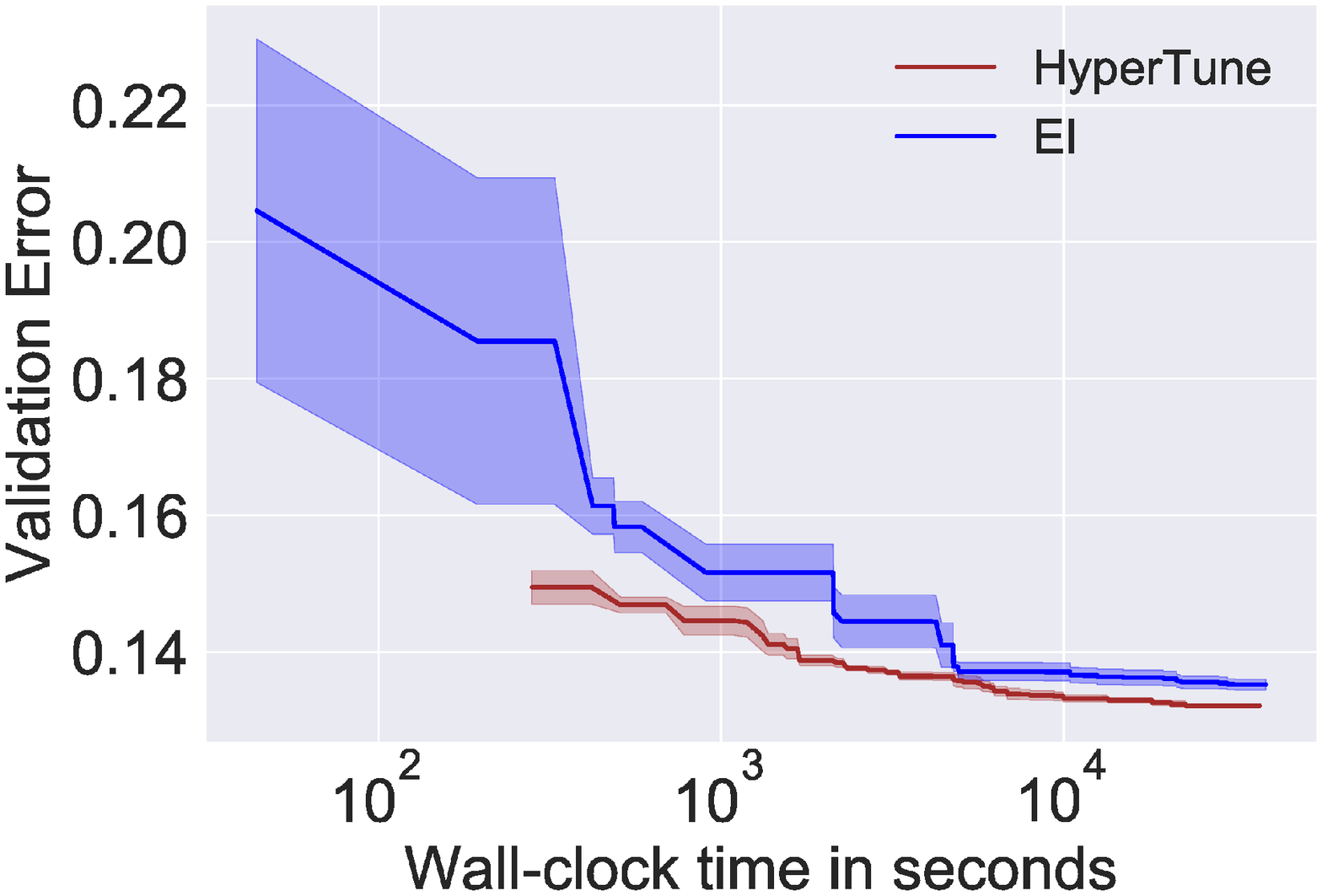}}\caption{EI and \emph{HyperTune} on validation dataset for SVM with kernel
approximation: (a) Letter, (b) MNIST, (c) Adult, and (d) Vehicle datasets.
\label{fig:SVMApproxValid}}
\end{figure*}
\begin{table*}
\centering{}%
\begin{tabular}{|c|c|c|c|c|}
\hline 
\multirow{2}{*}{Baselines} & \multicolumn{4}{c|}{Average of the test error $\pm$ standard error}\tabularnewline
\cline{2-5} 
 & Letter & MNIST & Adult & Vehicle\tabularnewline
\hline 
\hline 
Fabolas & {\small{}0.09$\pm$0.014} & {\small{}0.076$\pm$0.03} & \textbf{\small{}0.242$\pm$0.000} & {\small{}0.18$\pm$0.016}\tabularnewline
\hline 
Hyperband & {\small{}0.04$\pm$0.002} & {\small{}0.019$\pm$0.00} & {\small{}0.246$\pm$0.002} & \textbf{\small{}0.137$\pm$0.000}\tabularnewline
\hline 
\textit{HyperTune} & \textbf{\small{}0.03$\pm$0.000} & \textbf{\small{}0.018$\pm$0.00} & \textbf{\small{}0.242$\pm$0.000} & {\small{}0.138$\pm$0.000}\tabularnewline
\hline 
\end{tabular}\caption{Generalization performance on heldout dataset for SVM with kernel
approximation.\label{tab:svm_kern_test}}
\end{table*}

\subsection{Experimental Results}

\subsubsection*{Elastic Net}

Elastic net is a logistic regression classifier with $L_{1}$ and
$L_{2}$ penalty \cite{zou2005regularization}. The hyperparameters
are the ratio parameter that trades-off $L_{1}$ and  $L_{2}$ penalty,
and the $\alpha$ parameter that determines the magnitude of the penalty.
We tune the ratio parameter in the bound $[0,1]$. The penalty parameter
$\alpha$ is tuned in an exponent space of $[-7,-1]$. In \emph{HyperTune},
we assign a directional derivative sign $m=-1$ only for the hyperparameter
$\alpha$ as the complexity of the model decreases with the penalty
hyperparameter $\alpha$ while the other hyperparameter does not contribute
to model complexity. The results for \emph{HyperTune} and EI on
the validation dataset are shown in \ref{fig:elnet_valid}. \emph{HyperTune}
outperforms EI in all the datasets. Table \ref{tab:svm_test} reports
the generalization performance of all the methods on a heldout dataset.
\emph{HyperTune} performs better than Fabolas in all cases.

\subsubsection*{SVM with RBF Kernel }

In this experiment, we tune the cost parameter $C$ of SVM and the
length-scale $\gamma$ of RBF kernel. We tune both the hyperparameters
in an exponent space of $[-3,3]$. For \emph{HyperTune}, we assign
directional derivative signs $\mathbf{m}=\{1,1\}$ for both the hyperparameters
$C$ and $\gamma$. The complexity of the model increases with an
increase in both these hyperparameters. We compare the performance
of EI and \emph{HyperTune} on the validation dataset in Figure \ref{fig:svm_VALID}.
In all the datasets, \emph{HyperTune} significantly outperforms EI.\emph{}
We further record the generalization performance on the heldout dataset
in Table \ref{tab:svm_test}. The results show that \emph{HyperTune}
again performs better than Fabolas, particularly in Letter dataset
where the difference is significant.

\subsubsection*{SVM with Kernel Approximation using Random Fourier Features}

In our third experiment, we tune the hyperparameters of SVM with kernel
approximation using random Fourier features \cite{rahimi2007random}.
We tune the cost parameter $C$, length-scale of RBF kernel $\gamma$
and the number of Fourier feature bases in this experiment. Both $C$,
and $\gamma$ are tuned in the same bound as the previous experiment,
and the number of Fourier features are tuned in an exponent space
of $[3,12]$. We assign directional derivative signs $\mathbf{m}=\{1,1,1\}$
for all the hyperparameters since the model complexity increases with
these hyperparameters. In Figure \ref{fig:SVMApproxValid}, we plot
the results for \emph{HyperTune} and EI on the validation dataset.
\emph{HyperTune again} outperforms EI in all the datasets.  We also
report the performance of the baselines on the heldout dataset in
Table \ref{tab:svm_kern_test}. The results show that \emph{HyperTune}
is either compatible or better than Hyperband. Both \emph{HyperTune}
and Hyperband outperform Fabolas in all the datasets except Adult
dataset. Both Fabolas and \emph{HyperTune}, however, perform better
than Hyperband in Adult dataset.
\begin{figure}
\centering{}\includegraphics[width=0.4\paperwidth]{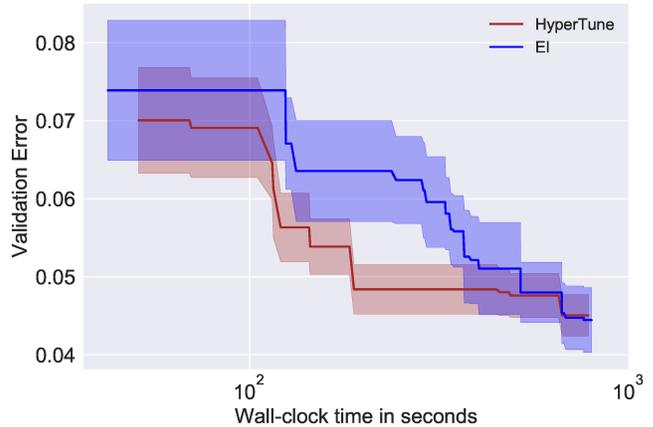}\caption{EI and \textit{HyperTune} on validation dataset for MLP.\label{fig:mlp_validError}}
\end{figure}
\begin{table}
\centering{}%
\begin{tabular}{|c|c|c|}
\hline 
\multirow{2}{*}{Baselines} & \multicolumn{2}{c|}{Average of the test error $\pm$ standard error}\tabularnewline
\cline{2-3} 
 & MLP on MNIST & CNN on CIFAR10\tabularnewline
\hline 
Fabolas & {\small{}0.09$\pm$0.008} & {\small{}0.28}\tabularnewline
\hline 
Hyperband & \textbf{\small{}0.04$\pm$0.003} & \textbf{0.21}\tabularnewline
\hline 
\textit{HyperTune} & {\small{}0.05$\pm$0.006} & \textbf{\small{}0.21}\tabularnewline
\hline 
\end{tabular}\caption{Generalization performance on heldout dataset for MLP and CNN. We
do not report the standard error for CNN as the experiment is conducted
only once.\label{tab:nn_test}}
\end{table}

\subsubsection*{Multi-layer Perceptron }

We tune five hyperparameters of an MLP on MNIST dataset. The hyperparameters
here are the number of neurons, dropout, mini-batch size, learning
rate, and momentum. We use stochastic gradient descent to learn the
model. In this experiment, we assign $m=1$ only for the number of
neurons in the hidden layers. We plot the results for EI and \emph{HyperTune}
on validation dataset in Figure \ref{fig:mlp_validError}. \emph{HyperTune}
performs better than EI on the validation dataset. We report the performance
of the baselines on the heldout dataset in Table \ref{tab:nn_test}.
Whilst Hyperband achieves the best performance in this experiment,
\emph{HyperTune} performs equally well. Both \emph{HyperTune} and
Hyperband perform better than Fabolas.

\subsubsection*{Convolutional Neural Network }

We tune six hyperparameters of a CNN on benchmark real-world CIFAR10
dataset \cite{krizhevsky2014cifar}. Using the benchmark architecture
configuration with 25 epochs, we tune batch size, dropout in the convolutional
layers, learning rate, momentum, number of neurons in the fully connected
layer, and dropout in the fully connected layer. For \emph{HyperTune},
we use $m=1$ only for the number of neurons in the fully connected
layer. We allocate a fixed a time budget of 24 hours to all the baselines
and the best performance on the heldout dataset is reported in Table
\ref{tab:nn_test}. Both Hyperband and \emph{HyperTune} achieve the
best generalization performance, and outperform Fabolas.\selectlanguage{british}%

\section{Conclusion\label{sec:conclusion}}

\selectlanguage{english}%
We have developed a fast hyperparameter tuning framework rooted in
the insights from PAC learning theory. We have identified a novel
way to leverage the trends in generalization performance from a subset
of the data to the whole dataset. We have effectuated this using directional
derivatives that signals the monotonic trends in the generalization
performance of the models for hyperparameters which directly dictates
model complexity. We evaluate the efficacy of our algorithm for tuning
the hyperparameters of several machine learning algorithms on benchmark
real-world datasets. The results demonstrate that our method is better
than Fabolas \cite{pmlr-v54-klein17a} and the generic Bayesian optimization.
We have also shown that HyperTune is compatible with the state-of-the-art
hyperparameter tuning algorithm, Hyperband \cite{li2016hyperband},
and is widely applicable.\selectlanguage{british}%

\bibliographystyle{aaai}
\bibliography{example}

\end{document}